\documentclass[10pt,twocolumn,letterpaper]{article}

\usepackage[pagenumbers]{cvpr} 

\usepackage{graphicx}
\usepackage{amsmath}
\usepackage{amssymb}
\usepackage{booktabs}
\usepackage{soul,color}
\usepackage{amsmath,bm}
\usepackage{subcaption}

\usepackage[pagebackref,breaklinks,colorlinks]{hyperref}

\usepackage[capitalize]{cleveref}
\crefname{section}{Sec.}{Secs.}
\Crefname{section}{Section}{Sections}
\Crefname{table}{Table}{Tables}
\crefname{table}{Tab.}{Tabs.}

\def\cvprPaperID{6} 
\def\confName{CVPR}
\def\confYear{2022}

\begin{document}

\title{Augmentation of Atmospheric Turbulence Effects\\on Thermal Adapted Object Detection Models}

\author{Engin Uzun\textsuperscript{1,2}\thanks{indicates equal contribution} \qquad\qquad Ahmet Anıl Dursun\textsuperscript{1}\footnotemark[1] \qquad\qquad Erdem Akagündüz\textsuperscript{2}\\
\textsuperscript{1}Dept. of Image Proc. \& Computer Vis. Technologies, ASELSAN Inc., Turkey\\
\textsuperscript{2}Graduate School of Informatics, Middle East Technical University, Turkey\\
{\tt\small \{enginuzun, aadursun\}@aselsan.com.tr}, {\tt\small akaerdem@metu.edu.tr}
}

\maketitle

\begin{abstract}
 Atmospheric turbulence has a degrading effect on the image quality of long-range observation systems. As a result of various elements such as temperature, wind velocity, humidity, etc., turbulence is characterized by random fluctuations in the refractive index of the atmosphere. It is a phenomenon that may occur in various imaging spectra such as the visible or the infrared bands. In this paper, we analyze the effects of atmospheric turbulence on object detection performance in thermal imagery. We use a geometric turbulence model to simulate turbulence effects on a medium-scale thermal image set, namely ``FLIR ADAS v2''. We apply thermal domain adaptation to state-of-the-art object detectors and propose a data augmentation strategy to increase the performance of object detectors which utilizes turbulent images in different severity levels as training data. Our results show that the proposed data augmentation strategy yields an increase in performance for both turbulent and non-turbulent thermal test images.
\end{abstract}
\section{Introduction}

\begin{figure}[t]
  \centering
   \includegraphics[trim=60 10 50 0,width=0.9\linewidth]{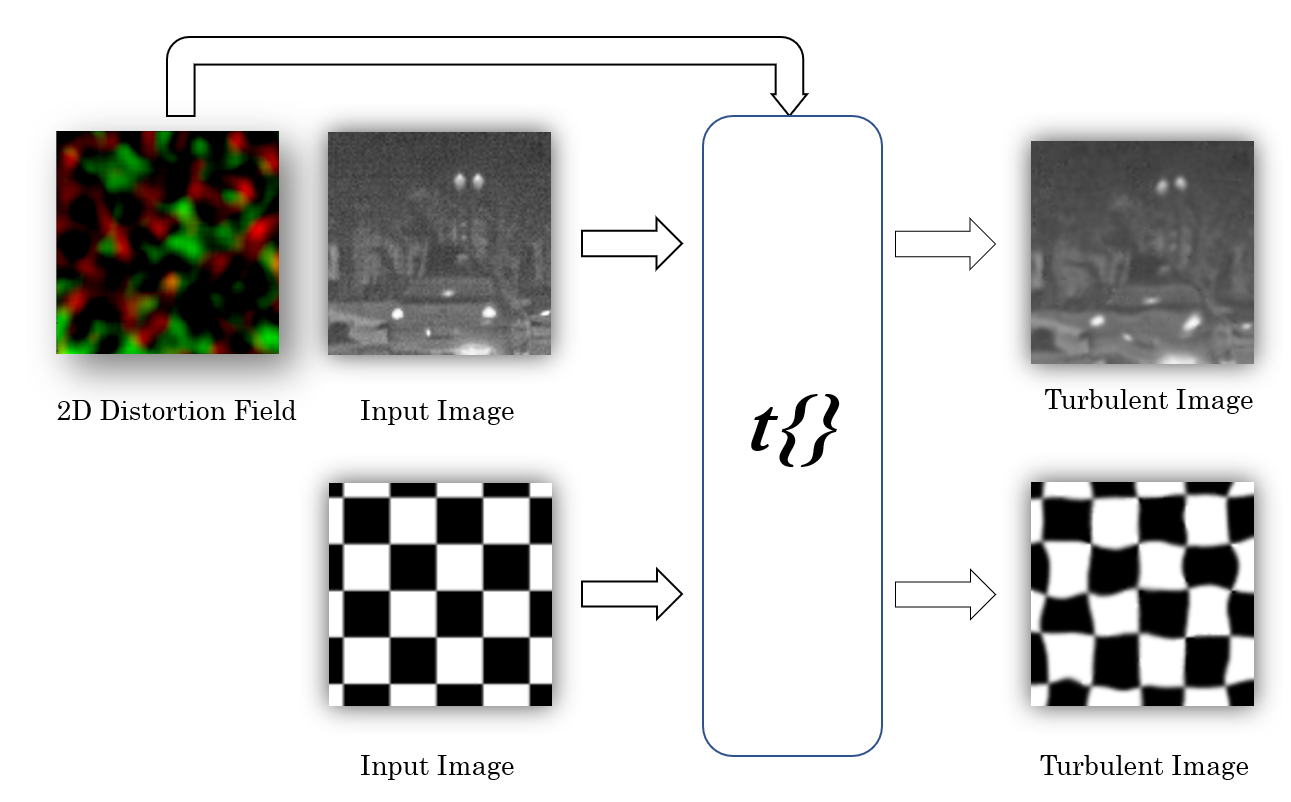}
   \caption{The turbulence generation processes on an IR and a chessboard images are illustrated. ${t\{\}}$ is the turbulence operation and  the vertical ($d_n^u(x,y)$ in red) and the horizontal ($d_n^v(x,y)$ in green) elements of the 2D distortion field are depicted in false color.}
   \label{figturbimgen}
\end{figure}





Caused by various atmospheric elements such as temperature, humidity or wind, atmospheric turbulence usually refers to the three-dimensional chaotic flow of air. Turbulence has a distorting effect on the propagation of light through air. It changes the refractive index of air in a spatio-temporal manner. Especially in long-range imaging systems, such distortions become apparent in various electromagnetic spectra. In astrophotography, the turbulence effect is assumed to be isoplanatic, meaning that the distortions are spatially non-varying. Adaptive optics is a common solution for mitigating the isoplanatic turbulence effects \cite{adaptiveoptics}. However, for ground-to-ground imaging systems, turbulence is anisoplanatic, which results in non-rigid distortions and blurring on the image. Such non-rigid distortions  possibly affect the performance of any computer vision algorithm, which is a fact that is not widely investigated in the literature. 

Theoretical studies on the effects of atmospheric turbulence in the infrared (IR) spectra can be found in the literature \cite{turbfriedcoh,turbth1,turbth2,turbth3}. For example \cite{turbfriedcoh} proposes the Fried coherence diameter as a measure of the optical transmission quality through the turbulent air, which reflects the relation of wavelength and imaging quality. According to this measure, as the wavelength increases, the resultant effect of turbulence tends to decrease. Therefore, the visible spectrum suffers more than the IR spectra from turbulence. However, the effect of turbulence on images also becomes significant in the IR spectra, when the refractive index structure parameter, ${C_n}^2$  \cite{turbfriedcoh,cn2}, and the path length are sufficiently high.

In this work, the effects of anisoplanatic turbulence on object detection are analyzed for IR imagery. Object detection is one of the trending challenges of computer vision and deep learning, which is being widely applied in several problem domains \cite{objdet1,objdet2}. The literature for object detection can be chronologically scrutinized in two mainstream approaches, namely the two-stage and the one-stage object detectors. One-stage approaches such as "YOLO" \cite{yoloBase} and its variants \cite{yolov2,yolov3,yolor,yolotolga},  are the successors of two-stage approaches and have lower complexities; hence, are popular for real-time applications. Studies on object detection in IR imagery are relatively new \cite{therObjDet1,therobjDet2,therobjDet3}. Similar architectures to visible domain are being utilized. However, due to relatively low-scales of IR image sets, detection success on IR imagery is lower when compared to visible band.  In order to tackle the data problem in IR imagery, recent studies focus on visible-to-IR domain adaptation and achieve satisfactory results using thermal-RGB pairs and supervised learning \cite{HerrmannIRAdapt,9636353}. Moreover, adversarial-based unsupervised methods are also utilized to thermal domain adaptation \cite{Akkaya_2021_CVPR}. Hence, thermal adaption is currently a de-facto practice in IR computer vision problems.  

Although several IR image and video sets are publicly available \cite{danaci2022survey}, to the best of our knowledge, there is no public IR image set that includes turbulence effects, for object detection or any other computer vision tasks. This is mainly because creating a set consisting of both turbulent and non-turbulent images of a given scene is a very expensive and difficult task. A feasible option is to create synthetic turbulence on existing image sets, using mathematical models of turbulence. The existing models in the literature \cite{turb3dsim2,turb3dsim3,turb3dsim1} generally represent the dynamics of atmospheric turbulence in 3D coordinates and project synthetic 3D scenes to 2D images. However, creating a large corpus of images with different 3D scene settings is also a very difficult task to achieve. Furthermore, in the thermal domain, creating 3D models with realistic material properties is problematic as well. Another approach can be working on existing thermal object detection datasets that additionally include 3D objects models. However, this case would require a reconstruction of the 3D scene with depth information, which is not available for almost any image set. 

Previous work on generating synthetic atmospheric turbulence on images mainly focuses on the visible spectrum. In \cite{turb2dsim4}, generative adversarial networks (GANs) are utilized for turbulent image generation in the visible spectrum, where sufficiently large-scale data can easily be collected. \cite{turb2dsim2} uses neural networks to simulate a geometric model on visible band images. In \cite{turb2dsim1,turb2dsim3}, physics-based models are proposed. There are recent studies that focus on the effects of atmospheric turbulence on long-range observation systems for limited applications fields \cite{10.1117/12.2177800}. Nevertheless, none of the aforementioned literature propose solutions to turbulence related issues of infrared computer vision problems. 

In this paper, we use a geometric model to create turbulence distortions on IR images. The model is operationalized on a set of IR images selected from the ``FLIR ADAS v2'' dataset \cite{flir}. Both the generated (turbulent) and original (non-turbulent) images are used so as to improve the performance of deep learning-based object detectors. In our experiments, we benchmark 3 different one-stage deep object detectors, which we transfer as pretrained and attempt to adapt to thermal domain via fine-tuning.  Moreover, we propose a data augmentation strategy that yields an increase in detection performance for both turbulent and non-turbulent thermal test images. The details of the geometric turbulence model, the benchmarked architectures, the experiments and our results are provided in the following sections.

\section{The Turbulence Model}

Previous studies that model the turbulence effect geometrically usually utilize a composition of blurring and random distortions \cite{turbmit1,turbmit2,turbmit3}. Blurring can be applied using basic low-pass filtering. Random distortions can be emulated using image warping and can be efficiently performed on GPUs \cite{warpgpu}. We utilize the following model for turbulent image generation using non-turbulent images:

\begin{multline}
F_n(x,y)=D((G_B(x,y) \circledast I_n(x,y)), ... \quad \quad \\
\quad  \quad \quad \quad \quad \quad \quad \quad \quad \quad d_n^u(x,y),d_n^v(x,y))
\label{eq1}
\end{multline}
where $F_n(x,y)$ is the source image, $D$ is the warping function, $\circledast$ is the convolution operation and $G_B(x,y)$ is a Gaussian kernel with variance $\sigma_B^2$, responsible for the blurring operation. Note that the warping operation is applied on both horizontal and vertical directions using the random distortion fields, $d_n^u(x,y)$ and $d_n^v(x,y)$, respectively, which are defined as:

\begin{equation}
d_n^u(x,y)=\gamma * (G_D(x,y) \circledast v_n^u(x,y))
\label{eq2}
\end{equation}
\begin{equation}
d_n^v(x,y)=\gamma * (G_D(x,y) \circledast v_n^v(x,y))
\label{eq3}
\end{equation}
where $\gamma$ is the amplitude of the random distortion and $G_D(x,y)$ is the Gaussian kernel with variance $\sigma_D^2$. $v_n^u(x,y)$ and $v_n^v(x,y)$ are random vectors with zero-mean, unit-variance normal distributions. Convolution operation with $G_D(x,y)$ provides spatial correlation of the random distortions over the image. $\sigma_D^2$ is used to adjust the strength of the spatial correlation while $\gamma$ is the amplitude of the distortions in the model. $\sigma_B^2$ is used to adjust the amount of blurring. In Fig. \ref{figturbimgen}, the proposed geometric turbulence model is depicted with a sample IR image and a chessboard image.

\subsection{Physical Approximation}

In order to relate the proposed geometric turbulence model parameters with real world turbulence conditions, in this section we formulate the relation between the magnitude of the pixel shift  (distortion vectors) and the model parameters, namely $\gamma$ and $\sigma_D^2$. Note that, we do not utilize $\sigma_B^2$ for this approximation since it is not related with the distortion process. Let $\bm{z(x,y)}$ be the random distortion vector over an arbitrary image region,

\begin{equation}
\bm{z(x,y)} = [d_n^u(x,y) ,d_n^v(x,y)]^T
\label{eq4}
\end{equation}
$\bm{z(x,y)}$ is a bivariate Gaussian distribution, where u and v components are independent of each other and are zero-mean. Variance of u and v components depends on the values of both $\gamma^2$ and $\sigma_D^2$. In Equations \ref{eq2} and \ref{eq3}, the vertical and horizontal distortions are derived from zero-mean, unit-variance Gaussian distributions. In the proposed model, these random variables, $v_n^u(x,y)$ and $v_n^v(x,y)$, are firstly filtered spatially by a 2D Gaussian kernel with $\sigma_D^2$ variance, then multiplied with the turbulence severity level, $\gamma$. Spatial filtering operation on $d_n^u(x,y)$ and $d_n^v(x,y)$ corresponds to the weighted linear combination of zero-mean unit-variance uncorrelated Gaussian random variables, where the weights are the parameters in $G_D(x,y)$. The equivalent variance of the spatially filtered random distortions are the sum of the squared kernel parameters. For a sufficiently large kernel, this summation is equivalent to the integration of the corresponding squared continuous bivariate Gaussian function in 2D space. Then, $\gamma$ is simply a multiplier over the variance. The resultant variance of $d_n^u(x,y)$ and $d_n^v(x,y)$, $\sigma_z^2$, can be expressed as,

\begin{equation}
\sigma_z^2 = \gamma^2\int \int G(p,\mu_p,\Sigma_p)^2 {dp} 
\label{eq5}
\end{equation}

where 
$\bm{\mu_p}$ is the mean and $\bm{\Sigma_p}$ is the covariance matrix of the bivariate Gaussian kernel. For our case, the Gaussian kernel $G_D(x,y)$ is assumed to be zero-mean and have constant diagonal covariance matrix $\bm{\Sigma_p}$. Given these conditions, in explicit form \ref{eq5} reduces to,



\begin{equation}
\sigma_z^2 \\ 
= \gamma^2\int \int (\dfrac{1}{2\pi\sigma_D^2}e^{-\dfrac{p_0^2 + p_1^2}{2\sigma_D^2}})^2 dp_0 dp_1
\label{eq8}
\end{equation}

The the resultant expression for $\sigma_z^2$ becomes,

\begin{equation}
\sigma_z^2= \dfrac{\gamma^2}{4\sigma_D^2\pi}
\label{eq9}
\end{equation}

Let $l(x,y)$ be the Euclidean norm of the random distortion vector $\bm{z(x,y)}$, which is given as,

\begin{equation}
l(x,y)= \sqrt{d_n^u(x,y)^2 + d_n^v(x,y)^2}
\label{eq10}
\end{equation}
$l(x,y)$ is also a random variable and because it is the Euclidean norm of a vector sampled from a normal distribution with zero-mean and constant diagonal covariance matrix, $l(x,y)$ is expected to have a Rayleigh distribution, which can be written as,

\begin{equation}
f_{ {l(x,y)}}({l(x,y)}) = \dfrac{l(x,y)}{\sigma_z^2}e^{-l(x,y)^2(2\sigma_z^2)}
\label{eq11}
\end{equation}

where $f_{ {l(x,y)}}({l(x,y)})$ is the probability density function of random variable $l(x,y)$. Mean of the $l(x,y)$, namely $\mu_{l}$, is the measure of how much the pixels are distorted on an image for the given parameters $\gamma$ and $\sigma_D^2$, and given as,

\begin{equation}
\mu_{l}= \sigma_z\sqrt{\dfrac{\pi}{2}}
= \dfrac{\gamma}{2\sqrt{2}\sigma_D}
\label{eq12}
\end{equation}

Note that, $\mu_{l}$ is a measure in pixel dimensions. In order to relate the distortions in the pixel dimensions to the real world measurements, a pin-hole camera model can be utilized using internal parameters of the thermal camera used in collecting the FLIR ADAS v2 dataset. Relation between pixel distortions and the corresponding real world distortion for both horizontal and vertical axes, $t_s^h$ and $t_s^v$ respectively, can be written as,

\begin{equation}
t_s^h = \dfrac{2\mu_{l}d_s tan(\alpha_h/2)}{W} = \dfrac{\gamma d_s tan(\alpha_h/2)}{\sqrt{2}\sigma_D W}
\label{eq13}
\end{equation}

\begin{equation}
t_s^v = \dfrac{2\mu_{l}d_s tan(\alpha_v/2)}{H}
= \dfrac{\gamma d_s tan(\alpha_v/2)}{\sqrt{2}\sigma_D H}
\label{eq14}
\end{equation}

where $d_s$ is the depth of the scene region, $\alpha_h$ and $\alpha_v$ are the horizontal and the vertical field of views respectively, $W$ is the horizontal pixel count and $H$ is the vertical pixel count of the camera. In \cite{flir}, horizontal and vertical fields of views of the thermal camera are given as $45^{\circ}$ and $37^{\circ}$ and the image resolution is given as 640x512. Then, \ref{eq13} and \ref{eq14} reduces to,

\begin{equation}
t_s^h = 0.0004576\dfrac{\gamma d_s}{\sigma_D} 
\label{eq15}
\end{equation}

\begin{equation}
t_s^v = 0.0004621\dfrac{\gamma d_s}{\sigma_D} 
\label{eq16}
\end{equation}

In the following sections, we will use the approximations given in \ref{eq15} and \ref{eq16} to determine the $\gamma$ levels to be used in our experimental setup.

\begin{figure*}[t]
\centering
\includegraphics*[width=1\textwidth]{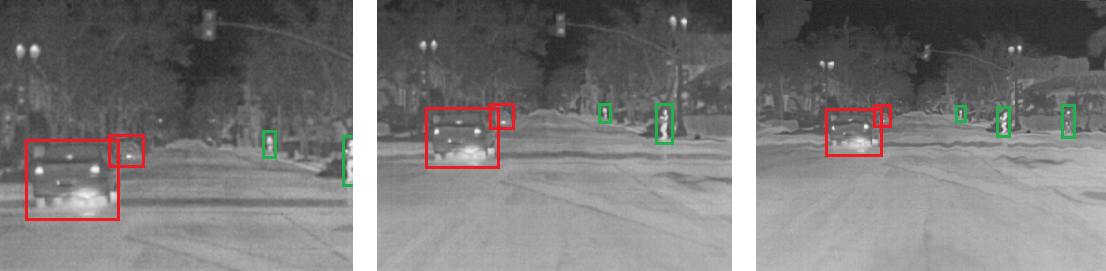}
\caption{The leftmost figure is a patch from a sample image of the FLIR ADAS v2  dataset. The middle and the rightmost figures are patches from the same image constructed with the proposed geometric turbulence model with $\gamma$ values 50 and 100, respectively. Annotations for the \emph{person} and \emph{car} classes are depicted in colored rectangles.}
 \label{fig:Montage}
\end{figure*} 

\section{Experimental Setup}
\subsection{The Image Set}
The experiments in this paper are implemented on the FLIR-ADAS v2 image set \cite{flir}. FLIR-ADAS v2 is a medium-scale image set annotated for object detection tasks in both thermal and visible bands. The set includes a total of 26,442 images with 15 different object classes. A total of 9,711 thermal and 9,233 RGB training/validation images are included with a 90\%-10\% train/validation split.  In this study, we use only the thermal images of the set, which are acquired with a Teledyne FLIR Tau 2. The resolution of the thermal images  is 640$\times$512 pixels. In our experiments, we utilize only the  \emph{car} and the \emph{person} classes because the number of samples from the other classes is insufficient for domain adaptation.

\subsection{Object Detection Architectures}
In our experiments, three state-of-the-art one-stage object detectors, namely VfNet\cite{zhang2020varifocalnet}, TOOD \cite{tood}, and YOLOR \cite{yolor}, are specifically chosen for their real-time performance. The reason that we deploy these three real-time models is because they are the state-of-the-art models that show leading detection performance in various common benchmarks. 

\subsubsection{VfNet}
Proposed by \cite{zhang2020varifocalnet}, VarifocalNet (VfNet) is an architecture that is designed on top of the fully convolutional one-stage object detector (FCOS) \cite{fcos}. FCOS originally proposes a pixel-wise approach similar to segmentation and utilizes a one-stage anchor free object detector. In \cite{atss}, the idea of adaptive training sample selection (ATSS) is developed on top of the FCOS architecture. 
VfNet is another improvement over this design family and has three main contributions to the baseline FCOS + ATSS approach. Firstly, VfNet proposes the so-called intersection over union-aware classification score (IACS) function, which is a scalar representation of a fusion of classification and the intersection-over-union (IoU) score. Secondly, they propose a novel loss function, namely Varifocal loss, which is designed to penalize their proposed IACS function. Varifocal loss function is based on the so-called Focal Loss \cite{lin2017focal} that is designed to handle the imbalance problem between positive and negative samples. 
Verifocal loss is utilized only for negative samples; whereas, for positive samples, Binary Cross-Entropy (BCE) Loss is employed with the multiplication of the ground truth class. Finally, they propose a star-shaped bounding box feature, which they claim to be more efficient for their IACS score function. In the original paper, VfNet is tested with various backbones, computation-wise lightest one of which is the ResNet-50 \cite{7780459}, as utilized in our experiments.

\subsubsection{TOOD}
Proposed by \cite{tood}, the Task-aligned One-stage Object Detection (TOOD) has two important contributions over the existing one-stage object detector models. First, instead of using parallel heads for classification and localization, TOOD utilizes a single, so-called task-aligned head in order to reduce any spatial misalignment. Second, TOOD uses a novel learning scheme, namely task alignment learning, which yields optimal anchor estimates for both classification and localization tasks. 
Although it is not one of the backbones tested for TOOD in the original paper, we again utilized ResNet-50 on this architecture, for a fair comparison with the VfNet.

\subsubsection{YOLOR}

YOLOR\cite{yolor} is basically a unified network architecture, which combines the explicit knowledge (i.e. shallow layer activations) and the implicit knowledge (i.e. deeper layer activations) of a network. Hence, more than an object detection specific architecture, YOLOR is designed as a multi-task learning model. However, the model is operationalized for object detection and performs comparable to leading non-real-time models such as the swin transformer \cite{9710580}, but achieves this in real-time. Because the model includes an encoder design of its own, no external backbone such as ResNet-50 is implemented for this model. In our experiments, we utilize the version called ``YOLOR-P6'', which is based on the ``YOLOv4-P6-light'' architecture.

\subsection{Training Configuration}
The experiments in this paper are deployed in paralel using 4 NVIDIA GeForce RTX 2080Ti GPUs. For all models, the batch size is set for 2 images for each GPUs. Open source MMDetection\cite{mmdetection} toolbox is utilized for training of VFNet and TOOD architectures. While training of these architectures, default hyper-parameters and original augmentation strategies are utilized. YOLOR-P6 architecture is trained using the official GitHub repository \cite{yolor}, which is provided by the authors. In a similar manner, default settings are utilized for training of YOLOR-P6. All experiments are run for 100 epochs. 

In order to augment the training and test sets with turbulent images of varying severity levels, we utilized four different values of $\gamma$ in our experiments, specifically 25, 50, 100 and 150. These different $\gamma$ values correspond to different atmospheric conditions, where the effect of the atmospheric turbulence increases directly proportional to $\gamma$ values.  We chose $\sigma_D^2$ and $\sigma_B^2$ as 25 and 1, respectively. In Figure \ref{fig:Montage}, the effects of applied turbulence on a sample image patch taken from the FLIR ADAS vs dataset are illustrated for $\gamma$ values 50 and 100. For instance, using Equation \ref{eq15}, for a scene with average depth of 100m, the selected four $\gamma$ values result in real world horizontal distortions of 0.23m, 0.46m, 0.92m and 1.37m. The vertical distortions can be calculated similarly using Equation \ref{eq16} and will result in close metric values. We believe that the selected $\gamma$ values span a realistic range of turbulence levels for real world applications. 

\section{Results}
\label{section:results}

\begin{table}[]
\centering
\resizebox{0.47\textwidth}{!}{%
\begin{tabular}{lllllllll}
\textbf{Aug} & \textbf{Model} & \textbf{Backbone} & \bm{$AP$} & \bm{$AP_{50}$} & \bm{$AP_{75}$} & \bm{$AP_{S}$} & \bm{$AP_{M}$} & \bm{$AP_{L}$} \\ \hline
\multicolumn{9}{c}{ \textbf{test set without turbulence}}  
\\ \hline
w/o  & VfNet             & ResNet -50   & 52.4 & 81.9 & 54.8 & 41.1 & 73.4 & 76.8 \\
w/   & VfNet             & ResNet -50   & 54.7 & 83.8 & 57.1 & 43.7 & 75.2 & 79.9 \\ \hline
w/o  & TOOD              & ResNet -50   & 46.1 & 74.6 & 47.3 & 33.1 & 71.1 & 76.3 \\
w/   & TOOD              & ResNet -50   & 46.8 & 75.4 & 47.9 & 33.7 & 71.3 & 77.8 \\ \hline
w/o  & YOLOR             & YOLOR-P6     & 53.1 & 80.5 & 56.2 & 40.6 & 77.3 & 82.9 \\
w/   & YOLOR             & YOLOR-P6     & 53.8 & 81.0 & 57.2 & 41.4 & 77.5 & 83.3 \\ \hline
\multicolumn{9}{c}{ \textbf{test set with turbulence} $\bm{\gamma = 25}$}    \\ \hline
w/o  & VfNet             & ResNet -50   & 51.0 & 80.4 & 53.1 & 39.2 & 73   & 77.0 \\
w/   & VfNet             & ResNet -50   & 53.9 & 83.1 & 56.1 & 42.7 & 74.7 & 79.4 \\ \hline
w/o  & TOOD              & ResNet -50   & 45.3 & 74.0 & 45.7 & 31.9 & 70.9 & 76.8 \\
w/   & TOOD              & ResNet -50   & 46.2 & 75.0 & 46.7 & 33.2 & 71.1 & 77.0 \\ \hline
w/o  & YOLOR             & YOLOR-P6     & 52.3 & 79.9 & 54.7 & 39.5 & 77.1 & 83.3 \\
w/   & YOLOR             & YOLOR-P6     & 53.1 & 80.5 & 56.0 & 40.6 & 77.2 & 83.8 \\ \hline
\multicolumn{9}{c}{ \textbf{test set with turbulence} $\bm{\gamma = 50}$}    \\ \hline
w/o  & VfNet             & ResNet -50   & 47.5 & 77.0 & 48.3 & 35.1 & 70.8 & 76.2 \\
w/   & VfNet             & ResNet -50   & 51.8 & 81.4 & 53.5 & 39.8 & 73.9 & 79.8 \\ \hline
w/o  & TOOD              & ResNet -50   & 43.4 & 71.9 & 44.0 & 29.7 & 69.7 & 77.0   \\
w/   & TOOD              & ResNet -50   & 44.9 & 73.9 & 45.1 & 31.5 & 69.8 & 77.2 \\ \hline
w/o  & YOLOR             & YOLOR-P6     & 50.0 & 77.8 & 51.8 & 36.5 & 75.7 & 82.8 \\
w/   & YOLOR             & YOLOR-P6     & 51.4 & 79.4 & 53.7 & 38.5 & 76.2 & 83.7 \\ \hline
\multicolumn{9}{c}{ \textbf{test set with turbulence} $\bm{\gamma = 100}$}   \\ \hline
w/o  & VfNet             & ResNet -50   & 36.7 & 64.4 & 36.1 & 23.4 & 61.3 & 72.5 \\
w/   & VfNet             & ResNet -50   & 45.6 & 76.0 & 45.1 & 32.3 & 69.9 & 78.1 \\ \hline
w/o  & TOOD              & ResNet -50   & 35.7 & 62.8 & 34.4 & 21.3 & 62.8 & 73.3 \\
w/   & TOOD              & ResNet -50   & 40.4 & 69.7 & 39.0 & 26.4 & 66.2 & 75.3 \\ \hline
w/o  & YOLOR             & YOLOR-P6     & 41.4 & 69.1 & 40.9 & 26.4 & 69.7 & 82.4 \\
w/   & YOLOR             & YOLOR-P6     & 45.9 & 74.5 & 46.3 & 31.8 & 72.5 & 82.6 \\ \hline
\multicolumn{9}{c}{ \textbf{test set with turbulence} $\bm{\gamma = 150}$}   \\ \hline
w/o  & VfNet             & ResNet -50   & 23.0 & 43.8 & 21.0 & 12.1 & 43.7 & 61.1 \\
w/   & VfNet             & ResNet -50   & 38.9 & 68.6 & 36.5 & 25.0 & 64.2 & 76.2 \\ \hline
w/o  & TOOD              & ResNet -50   & 26.0 & 49.3 & 24.1 & 13.0 & 49.4 & 68.1 \\
w/   & TOOD              & ResNet -50   & 35.4 & 63.7 & 33.2 & 20.9 & 61.6 & 74.1 \\ \hline
w/o  & YOLOR             & YOLOR-P6     & 31.1 & 55.8 & 29.9 & 16.2 & 58.2 & 78.6 \\
w/   & YOLOR             & YOLOR-P6     & 39.7 & 67.9 & 38.4 & 24.5 & 67.4 & 81.3 \\ \hline
\end{tabular}%
}
\caption{Mean Average Precision results obtained for different experiments with different models, with or without turbulent image augmentation for varying levels of turbulence gain $\gamma$.}
\label{tab:resultsTable}
\end{table}

The results of all the experiments are presented as average precision (AP) values in Table \ref{tab:resultsTable}. In this table, each row represents a separate experiment. The first column, namely \emph{Aug}, indicates if turbulent data is augmented during training (w/) or not (w/o). The following two columns state the \emph{Model} and the \emph{Backbone} used. The \emph{AP} column is the mean average precision calculated over the two categories; whereas \emph{AP\textsubscript{50}} and \emph{AP\textsubscript{75}}  represent the cases when IoU is selected as 0.5 and 0.75, respectively. The rightmost three columns set forth the AP values for the test subsets, where the test samples are distributed according to their pixel size. \emph{AP\textsubscript{S}} is the performance score for the subset that includes test objects smaller than 32$\times$32 pixels; whereas when calculating \emph{AP\textsubscript{L}}, test objects larger than 96$\times$96 pixels are used. \emph{AP\textsubscript{M}} represents the performance of the test objects that have sizes in between.

\begin{figure*}[t]
\begin{subfigure}{.5\textwidth}
  \centering
  \includegraphics[trim={40 0 50 0}, clip, width=1\linewidth]{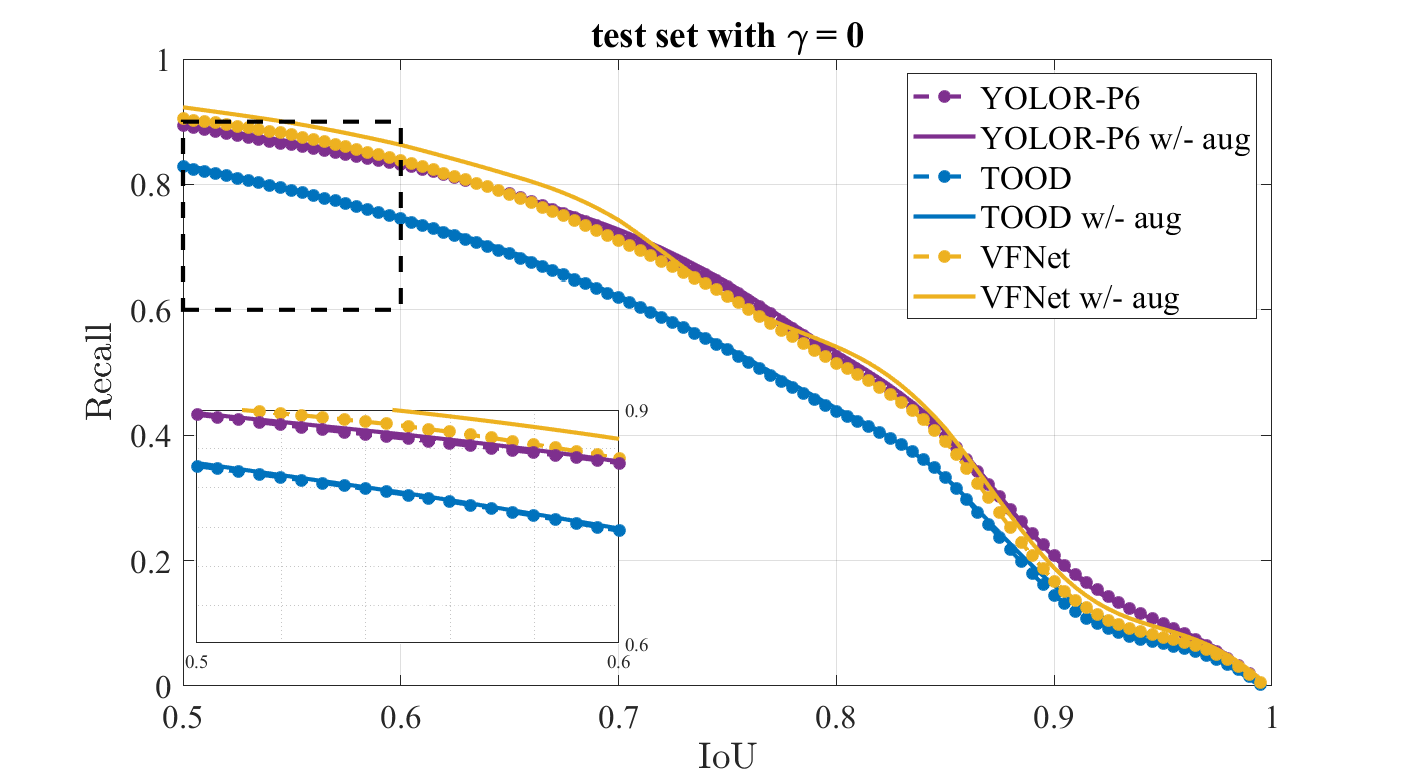}  
\caption{}
  \label{fig:iou_recall_a}
\end{subfigure}
\begin{subfigure}{0.5\textwidth}
  \centering
  \includegraphics[trim={40 0 50 0}, clip, width=1\linewidth]{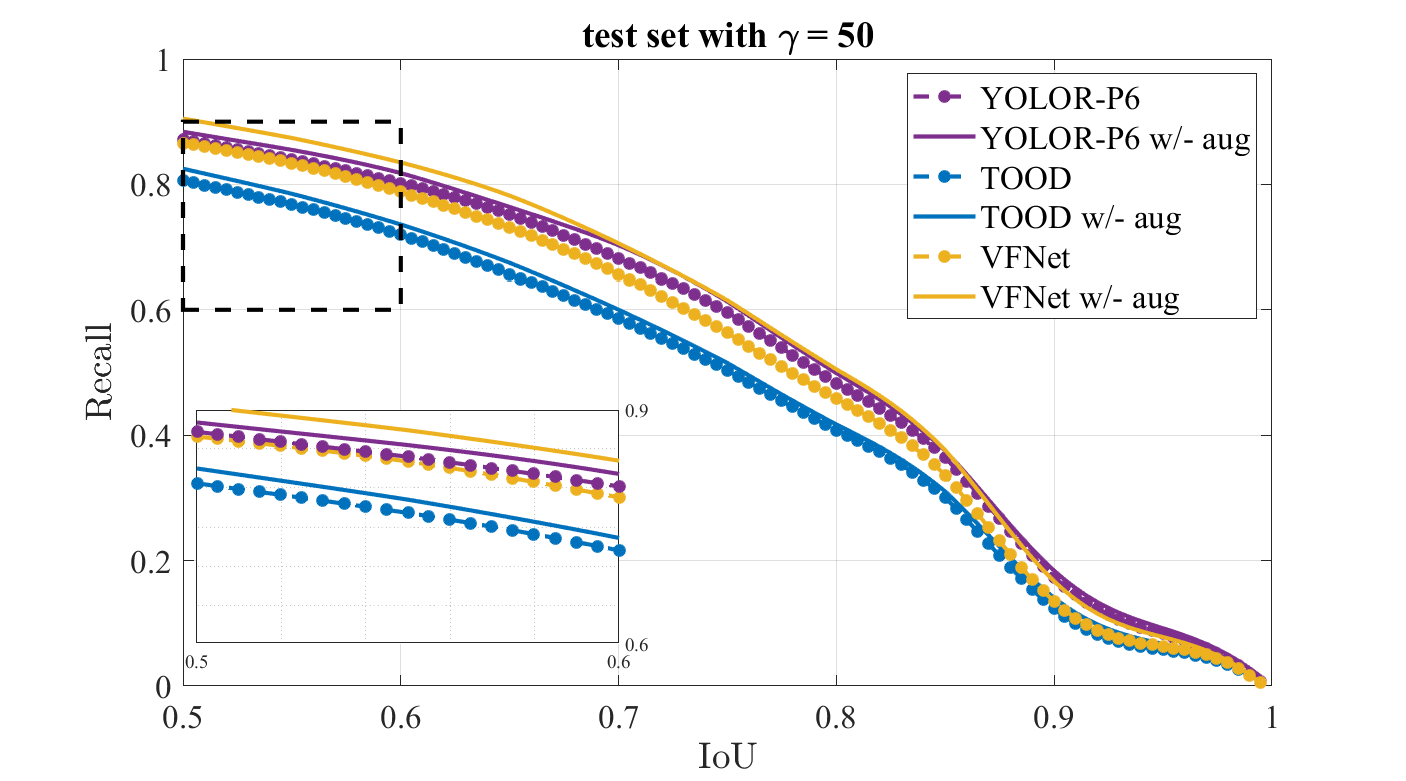}   
 \caption{}
  \label{fig:iou_recall_b}
\end{subfigure}
\begin{subfigure}{0.5\textwidth}
  \centering
  \includegraphics[trim={40 0 50 0}, clip, width=1\linewidth]{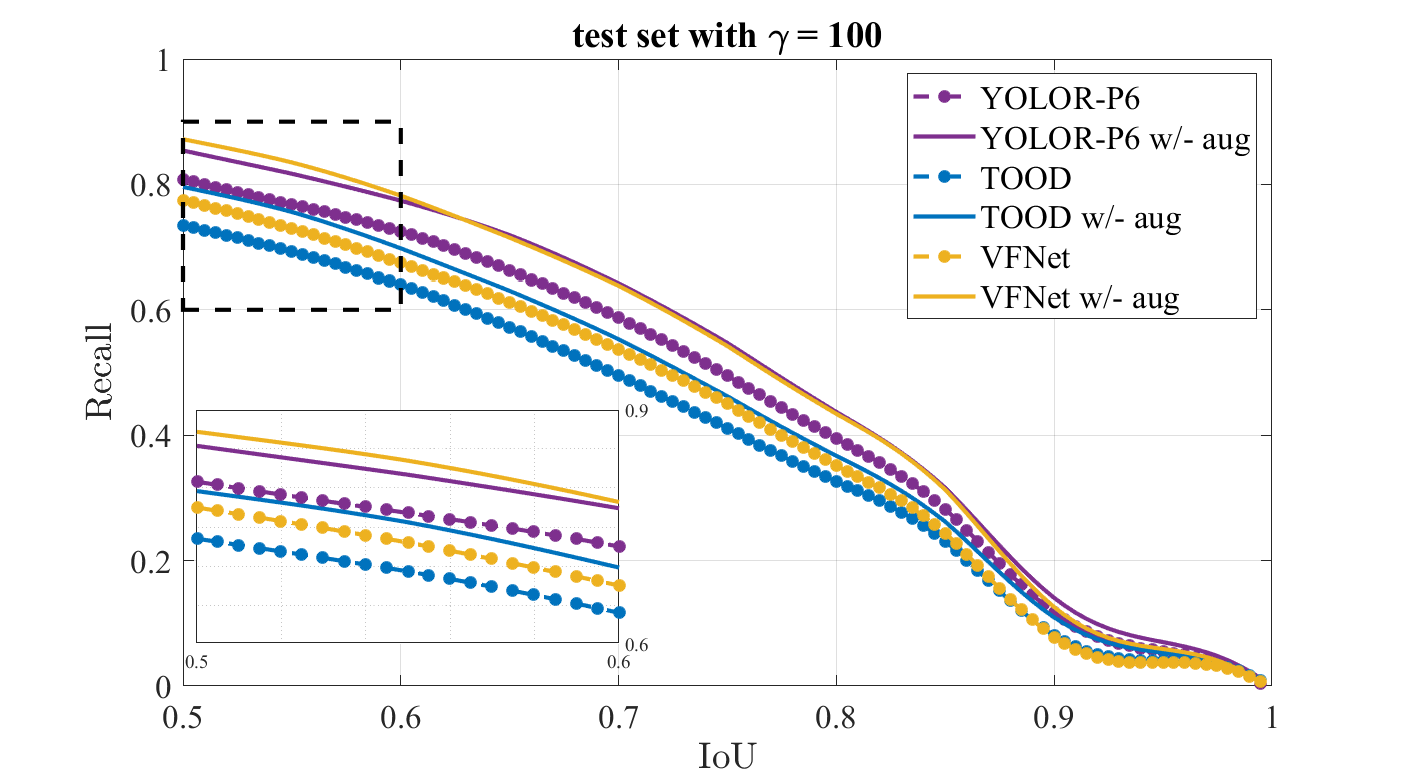}   
 \caption{}
  \label{fig:iou_recall_c}
\end{subfigure}
\begin{subfigure}{0.5\textwidth}
  \centering
  \includegraphics[trim={40 0 50 0}, clip, width=1\linewidth]{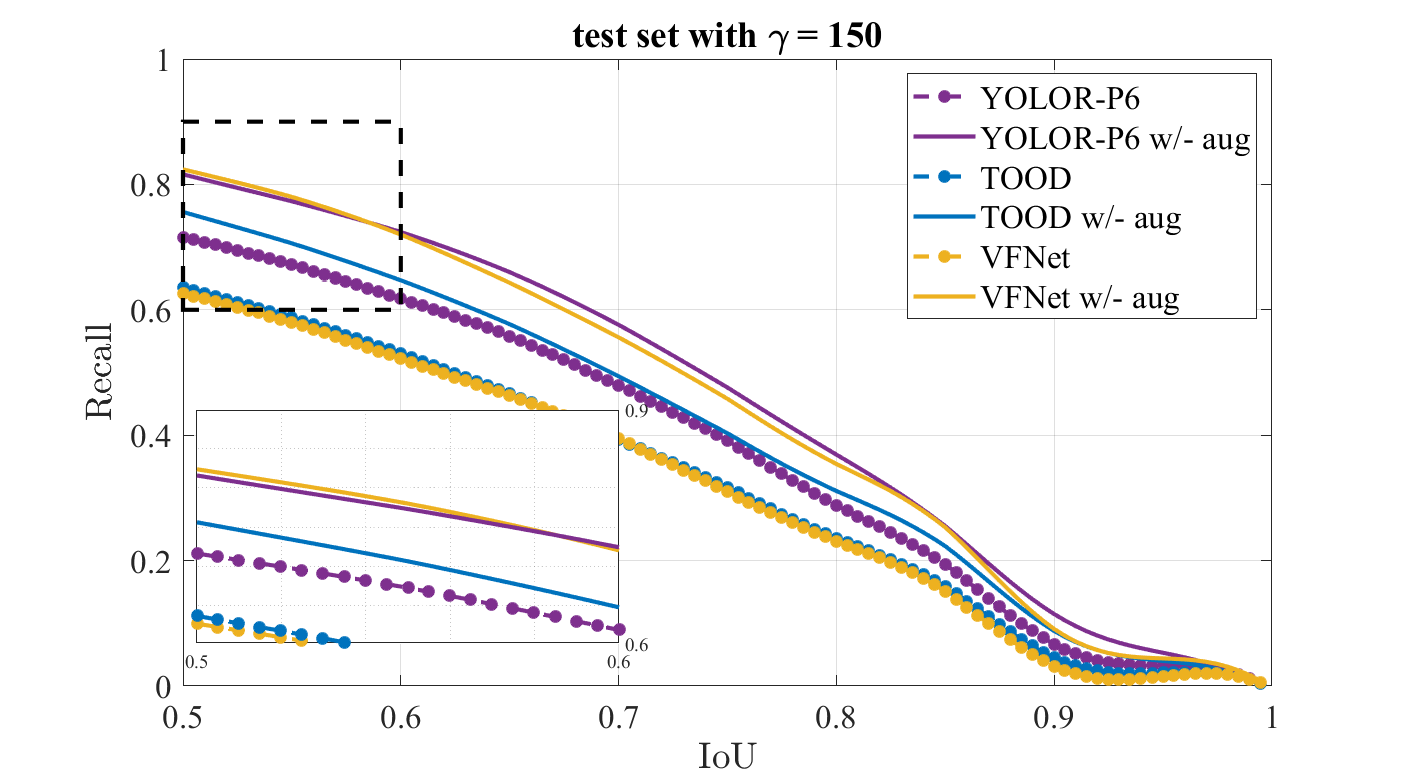}   
\caption{}
  \label{fig:iou_recall_d}
\end{subfigure}
\caption{``IoU vs Recall'' curves for all models, with and without turbulent data augmentation, for selected $\gamma$ levels.}
\label{fig:IvsR}
\end{figure*}

A rigorous analysis of Table \ref{tab:resultsTable} provides several important implications. To begin with, for all models and for all turbulence levels of the test sets, augmenting turbulent images of different $\gamma$ values during training always increase detection performance. This is one very important result of our experiments showing that the proposed augmentation strategy can be generalized for any thermal adaptation or model training experiment. Another important observation is that as the turbulence level of the test set increase, the positive effect of the proposed augmentation strategy also inevitably increase. We believe that this is an indicator of the robustness of the proposed augmentation strategy.

When we analyse the individual behaviour of the models, different characteristics can be observed. Although VfNet exhibits slightly better performance for the test sets with lower levels of turbulence, we see that as the turbulence level of the test sets increases, YOLOR becomes the leading model in general. However, the difference in performance between the VfNet and YOLOR models can be considered negligible. On the other hand, TOOD performs poorer than the other two models in general, which is consistent with its relative performance on RGB, i.e. before thermal adaptation. The only consistent behaviour for all models is that regardless of their contrasting performance when trained without turbulent data augmentation, our proposed augmentation strategy brings the models to comparable performance. This can be observed for the experiments with the highest turbulence levels of test sets, when VfNet performance becomes comparable to YOLOR as a result of our augmentation strategy, following a dramatic increase of mean \emph{AP} value of 23.0 to 38.9.

The \emph{AP\textsubscript{S}} values in Table \ref{tab:resultsTable} show that for small objects, detection performance falls dramatically at high levels of turbulence. This is not a surprising fact, because as the object is farther from the camera, not only its pixel size decreases, but the effect of turbulence increases as well. Nonetheless, Table \ref{tab:resultsTable} clearly shows that the impact of the proposed augmentation strategy is the strongest when the object size is small and the turbulence levels are higher. The reader should note that, before their adaptation, the benchmarked models are already trained with various augmentation strategies, which do not help them to overcome high-level turbulence effects. For medium and large objects, the positive effect of the augmentation strategy is minimal if the turbulence levels are low. However, for none of the cases, it is diminishing, i.e. \emph{AP\textsubscript{M}} and \emph{AP\textsubscript{L}} values are always higher for experiments where turbulent image augmentation is utilized.  

In order to examine the the localization success of each experiment, \emph{AP\textsubscript{50}} and \emph{AP\textsubscript{75}} columns are provided in Table \ref{tab:resultsTable}. Furthermore, ``IoU vs Recall'' graphs are given in Figure \ref{fig:IvsR}. In this figure, for different levels of $\gamma$, recall curves with respect to varying IoU are depicted, for all models, with or without turbulent data augmentation. Both the \emph{AP\textsubscript{75}} values in Table \ref{tab:resultsTable} and recall curves in Figure \ref{fig:IvsR} show that YOLOR model provides finer localization. In addition, we see from Figures \ref{fig:IvsR}a, \ref{fig:IvsR}b, \ref{fig:IvsR}c and \ref{fig:IvsR}d that as the turbulence levels increase, the impact of the proposed augmentation strategy becomes dramatically effective for any level of localization (i.e. IoU).


\section{Conclusions and Future Directions}
We propose a data augmentation strategy to increase the performance of thermal-adapted, real-time, one-stage object detectors under varying atmospheric turbulence conditions. In order to create turbulent images, we use a computation-friendly geometric turbulence model that can easily be implemented for online learning systems. Our results show that for all turbulence levels of the test sets, including the experiment, where there is no turbulent images in the test set, augmenting turbulent images of different severity levels always increases the detection performance for all models. Especially when the turbulence levels are high, the impact of the proposed augmentation scheme becomes much clear, such that it brings different model performances to a similar satisfactory level. 

The geometric distortions created by atmospheric turbulence usually demonstrate an unusual character in images. Conventional augmentation methods, such as affine transformations, do not have the capacity to simulate the effects of atmospheric turbulence, especially for long-range vision problems and small objects. We believe that turbulent image augmentation has the capacity to become a de facto practice for visible and thermal vision systems that utilize deep learning models.

In this paper, atmospheric turbulence effects are augmented using a geometric model for only the object detection problem. We believe that the approach can be expanded to other deep learning-based solutions of several IR vision problems. Furthermore, although using a computation-friendly geometric turbulence model such as ours has its advantages in practice, by using physics-based turbulence models and collecting data from calibrated scenes, more realistic turbulent images can be obtained. Such an effort would necessarily increase the impact of the proposed augmentation strategy.


{\small
\bibliographystyle{ieee_fullname}
\bibliography{egbib}
}

\end{document}